# Multiple Realizability and the Rise of Deep Learning

**Sam Whitman McGrath (sam_mcgrath1@brown.edu)**
Department of Philosophy
Department of Cognitive and Psychological Sciences
Brown University

**Jacob Russin (jake_russin@brown.edu)**
Department of Computer Science
Department of Cognitive and Psychological Sciences
Brown University

**Abstract**

The multiple realizability thesis holds that psychological states may be implemented in a diversity of physical systems. The deep learning revolution seems to be bringing this possibility to life, offering the most plausible examples of man-made realizations of sophisticated cognitive functions to date. This paper explores the implications of deep learning models for the multiple realizability thesis. Among other things, it challenges the widely held view that multiple realizability entails that the study of the mind can and must be pursued independently of the study of its implementation in the brain or in artificial analogues. Although its central contribution is philosophical, the paper has substantial methodological upshots for contemporary cognitive science, suggesting that deep neural networks may play a crucial role in formulating and evaluating hypotheses about cognition, even if they are interpreted as implementation-level models. In the age of deep learning, multiple realizability possesses a renewed significance.

## Introduction

Over the last decade, astonishing advances have taken place in artificial intelligence (AI), with state-of-the-art models equaling or surpassing humans on many tasks in domains such as natural language processing (Brown et al., 2020; Bubeck et al., 2023), vision (Croitoru et al., 2023), and strategic game-playing (Mnih et al., 2015; Silver et al., 2016; Silver et al., 2018). These breakthroughs have been powered by deep learning, an approach to AI involving the use of deep neural networks (DNNs) (LeCun, Bengio & Hinton, 2015; Goodfellow, Bengio & Courville, 2016).

Though some have argued that DNNs are, at best, merely mimicking human capacities (Bender & Koller, 2020; Chomsky et al., 2023), they have proved to be highly predictive of human neural responses in multiple cognitive domains (Yamins et al., 2014; Goldstein et al., 2022). Recent work has also shown that linguistic constructs used to explain certain human competencies, like syntactic parse trees, can be reconstructed from the internal states internal states of Large Language Models (LLMs; Manning et al., 2020), and seem to be causally implicated in their grammatical behavior (Chen et al., 2023). Such findings suggest that the cognitive capacities of these models may be more human-like than naïve assessment would suggest (McGrath, Russin, Pavlick, Feiman, 2023; Millière, Forthcoming).

Recent years have also seen renewed philosophical debate about the 'multiple realizability' of the mind, the possibility that the same psychological states found in humans might be realized in multiple kinds of physical systems. According to the multiple realizability thesis, "there might be non-carbon-based or non-protein-based biological organisms with mentality, and we cannot *a priori* preclude the possibility that nonbiological electromechanical systems, like the 'intelligent' robots and androids in science fiction, might be capable of having beliefs, desires, and even sensations" (Kim, 2006). DNNs may be the breakthrough that transforms this speculative possibility into an empirical reality.

In this paper, we explore the implications of the rise of DNNs for the multiple realizability thesis, which has raised issues at the intersection of philosophy, psychology, and artificial intelligence since its introduction (Putnam, 1967). Our central focus is the claim that multiple realizability implies that the study of the mind is independent and autonomous from the study of its realization in the brain (or in an artificial analogue). We argue that this implication does not hold—one can accept the possibility of multiple realization, while embracing neuroscience and contemporary DNNs as key sources of insight into cognition.

Though seldom articulated explicitly, the inference from multiple realizability to this form of methodological autonomy has played an important role in shaping research in this area, and rejecting it has key upshots for contemporary cognitive science. Here, we draw out these upshots by showing how two highly influential articles from opposed philosophical camps rely on this mistaken inference.

Fodor and Pylyshyn (1988) relies on it to insulate cognitive science from the 'connectionist' neural network research prominent at the time (Rumelhart et al., 1987). Fodor and Pylyshyn's objection is still invoked today to undercut the significance of modern neural networks (Quilty-Dunn, Porot & Mandelbaum, 2023). Rejecting it opens the door for contemporary DNNs to contribute to the formulation and evaluation of psychological hypotheses.

Bechtel and Mundale (1999) puts forward a very different argument, the goal of which is to reject multiple realization and defend the reduction of psychological states to brain states. Though diametrically opposed to Fodor's conclusions, Bechtel and Mundale's argument relies on the same alleged connection between multiple realization and methodological autonomy. As such, it falls to the same objection.

Throughout the majority of this paper, we will be focusing on what does and does not follow from the multiple realizability thesis. In the concluding discussion, we take a step back and examine how the success of deep learning bears on the truth of multiple realizability and methodological autonomy more directly.

## Multiple Realizability and the Autonomy of Cognitive Science

The multiple realizability thesis asserts that mental states and processes, like forming a belief or remembering an event, can be realized or implemented in a range of different physical systems. As even its staunchest opponents acknowledge (Polger & Shapiro, 2023), multiple realizability has attained the rarest of distinctions within philosophy—it is a source of widespread agreement among contemporary analytic philosophers.

The multiple realiza*tion* thesis goes further, asserting that this is not just an *a priori* metaphysical possibility, but a plausible empirical reality. Defenders of multiple realization may argue from general empirical facts, appealing to phenomena like convergent evolution or neural plasticity (Putnam, 1967), or develop specific cross-species case studies (Mallatt & Feinberg, 2021; Strappini, Martelli, Cozzo & di Pace, 2020). Such cases have proved to be controversial (Polger & Shapiro, 2016). Here, we will consider both multiple realizability and realization when each is relevant.

Much of the consensus around multiple realizability can be credited to the intuitive force of thought experiments involving alien and artificial intelligences. From the outset, however, this thesis was tightly bound to functionalism, a theory which defines mental states via their functional roles, rather than their physical constitution, and inspired by reflection on the Turing Machine (Putnam, 1960; Putnam, 1967; Putnam, 1973). Before Turing, reducing psychological states to brain states seemed to be the only way to locate the mind within a physical world without invoking a mysterious Cartesian "ghost in the machine" (Ryle, 2000). The advent of modern computing fundamentally altered this state of affairs by demonstrating that a physical system with the right kind of functional organization could replicate key aspects of human intelligence, like truth-preserving logical inference (Fodor, 1975). On this new 'computer model of the mind,' mental states are the 'software' and the brain is the 'hardware' on which they run. But just as a computer program can be implemented in different kinds of hardware, so too the mind might be realized not only in flesh-and-blood brains, but in silicon and perhaps even other undiscovered alien substances. These new philosophical views proved immensely influential, promising to make psychological states respectable objects of scientific study without reducing them to brain states or behavioral dispositions.

Multiple realizability is typically credited with securing the autonomy of higher-level cognitive sciences, like psychology and linguistics, from neuroscience and biology (Fodor, 1974). It is crucial, however, to distinguish two very different kinds of autonomy that often get conflated.

The first sense of autonomy is *ontological*. Mind-brain type identity theory requires a direct identification of types of psychological states with types of brain states, a one-to-one mapping between descriptions of the mind and descriptions of the brain. If psychological states are multiply realizable, however, this relation is one-to-many: for a given functional state, there are many ways it can be physically realized. This means that types of mental states are not natural kinds at the implementation level. Multiple realizability thereby provides a direct argument for the irreducibility of psychological constructs and, in so doing, secures an autonomous domain of entities for cognitive scientists to study.

The second sense of autonomy is *methodological*, rather than ontological. It asserts that the study of the mind can and must be pursued independently of the study of its implementation in the brain (Coltheart, 2004; for discussion, see McGeer, 2007). Put bluntly, methodological autonomy alleges that *neuroscience is irrelevant* to theorists working at the cognitive or computational level (Marr, 1982). This goes beyond the assertion that there are irreducible psychological constructs and puts forward a further methodological claim about how to go about formulating, verifying, and falsifying hypotheses about them.

If the human brain is just one of a diverse range of physical realizations of the mind's computations, the intuition goes, researchers interested in these computations should not get distracted by the local, contingent, implementation-level facts uncovered by neuroscientists. After all, in an alternate realization, these implementation-level details may look quite different. The cognitive scientist is essentially a *software reverse-engineer* and, as with most programmers, need not concern themselves with the hardware. Of course, whatever constructs cognitive scientists settle on must be implemented in the brain *somehow*, but this question can be postponed or passed off to a different department (Chomsky, 1968). Lawrence Shapiro calls this kind of methodological autonomy functionalism's "great promise"—"its suggestion that it is possible to investigate minds at a level of description that makes unnecessary an appreciation of the bodies and brains in which they are implemented" (Shapiro, 2004).

## Distinguishing Between Methodological Autonomy and Ontological Autonomy

Though often mentioned in the same breath, methodological autonomy and ontological autonomy do not stand on equal footing. The inference from multiple realizability to ontological autonomy is clear. Defenders of identity theory have argued that weaker forms of multiple realization may be compatible with their position (Polger, 2002; Polger, 2006), but as long as we grant some plausible premises—for example, that psychological kinds cannot be identified with wildly disjunctive physical kinds (Fodor, 1974)—standard multiple realization does preclude the possibility of identifying mental states with brain states.

More generally, what multiple realizability makes salient is the added explanatory value that psychological constructs contribute not despite, but *because* they provide a higher, more abstract level of description. Abstracting away from the messy details of particular realizations allows us to identify regularities and formulate explanatory generalizations that may not be readily apparent at the lower level (Putnam, 1973; Pylyshyn, 1984). Within a broadly Quinean, naturalistic framework (Quine, 1948; Quine, 1981), explanatory indispensability is taken to be the best guide to ontological commitment. As such, the ontological autonomy of psychological constructs seems secure.

Methodological autonomy is a different story. The fact that there are explanatory generalizations involving mental constructs does not directly entail anything about what the process of discovering them will look like. Adding that these constructs are or could be multiply realized does not change this fact. Whether knowledge of the brain is of use in understanding the mind is not a matter of how many alternate realizations the mind might have. It is a matter of whether the brain (or an artificial analogue) proves to be interpretable in higher-level terms—whether we can extract functionally relevant characterizations from an understanding of this particular physical realization (McGrath, Russin, Pavlick & Feiman, 2023). When we can do so, the fact that this functional organization might be reflected differently in different physical implementations is inconsequential.

The inference from multiple realizability to methodological autonomy seems to rely on the principle that, if the specific physical substrate is inessential to the mind, then the *study* of this specific physical substrate must be inessential to the *study* of the mind. But this principle simply does not hold.

To drive the point home, consider the game of chess (McGrath, Russin, Pavlick & Feiman, 2023). Chess is multiply realizable, and multiply realized. You can play over a traditional board, or with coconuts in the sand, or run a chess program on a computer. What it is to be a pawn, say, cannot be identified with or reduced to any specific properties of one of these realizations—pieces must be individuated at a higher-level of abstraction, specifying their legal moves and starting position. As such, chess pieces possess just the kind of ontological autonomy that proponents of multiple realizability ascribe to psychological constructs.

This analogy may seem to vindicate the inference to methodological autonomy. After all, for someone studying chess strategy—someone interested in the principles of quality pawn play, say—physical facts about how the pawns are implemented are irrelevant. Using your pawns to establish control of the center of the board is good strategy, whether they are plastic pieces or icons on a computer screen.

However, what underlies this intuition is not multiple realizability, but the fact that *we already know the rules of chess in advance*. Attending to the physical details of a particular realization cannot teach us anything new simply because there is *nothing left to learn*.

If this were not the case, however—if we did not know the rules and needed to infer them as we watched people play—then lower-level facts *could* supply crucial clues. For example, the fact that the pieces in the front row are all the same *size* and *shape* would be a key indicator that they all move the same way and form a unified functional kind: pawns. Again, things could be different—when playing with coconuts, such uniformity is likely to be unattainable. But the possibility of inscrutable or uninformative implementations does not detract from the evidential value of implementation-level details that do provide clues to function.

As cognitive scientists studying the mind, we are like the observer trying to reconstruct rules of chess by watching people play. We are not provided knowledge of the rules in advance but must reverse engineer them through careful observation and intervention. Facts about the brain form a key part of this inferential base and, as such, have an important role to play in the formulation and evaluation of psychological hypotheses. Neither multiple realizability nor multiple realization provides a principled reason to reject this explanatory relevance—one can embrace these theses without trying to insulate psychology from neuroscientific findings. If this strong form of methodological autonomy is to be defended, it will require a different argument.

## Fodor, Pylyshyn, and the "Mere Implementation" Objection

Though methodological autonomy has been discussed most explicitly in relation to the study of the brain, it is equally relevant to debates about the theoretical significance of artificial neural networks. In Fodor and Pylyshyn's influential 1988 article, "Connectionism and Cognitive Architecture: a Critical Analysis," they argue that neural networks offer, at best, mere implementations of classical symbolic computational architectures. As such, no matter how impressive, complex, and human-like these models' behavior becomes, they will remain irrelevant to theorists interested in cognitive characterizations of the mind. The possibility of mere implementation is still invoked by defenders of similar views today (Quilty-Dunn et al., 2023). As contemporary DNNs have blown past prior limitations, fulfilling the potential that connectionists identified in the 1980s, assessment of their theoretical significance and the cogency of this objection has become more pressing.

To begin, it is important to acknowledge what this critique gets right. Many of the connectionists that Fodor and Pylyshyn were attacking were *eliminativists* who held that the success of artificial neural networks warranted the wholesale replacement of classical explanatory constructs, like the Language of Thought and compositional syntactic structures (Fodor, 1975; McClelland et al., 1987; Rumelhart et al., 1987). The success of today's DNNs has prompted similar calls (Piantadosi, 2023).

Against an eliminativist opponent, the mere implementation objection makes an important contribution. Multi-layer neural networks can, in principle, approximate any function (Cybenkot, 1989), including those specified by a classical symbolic account. Despite important differences from classical models, then, it is a genuine empirical possibility that the success of contemporary DNNs is a result of their realizing some of the same explanatory constructs and operations that psychologists have long invoked. If this is the case, eliminativism is clearly unwarranted. If the ability to generate human-like sentences displayed by Large Language Models (Brown et al., 2019), for example, is a byproduct of their implementation of compositional syntactic structure—as has been suggested by recent findings (Manning et al., 2020; Tenney et al., 2019)—then LLMs' success does not repudiate decades of syntactic theory, but provides a striking confirmation of it (though it may still transform our understanding of how humans acquire knowledge of syntax).

However, this useful contribution is overshadowed by the way in which the claim that neural networks are implementation-level models gets misused to dismiss these networks as uninformative for theorists working at the cognitive level. It is here that the inference from multiple realization to methodological autonomy comes into play. If Fodor and Pylyshyn can position artificial neural networks as implementation-level models, and multiple realizability implies that implementation-level details are irrelevant to higher-level inquiry, then it would follow that these models need not impact psychologists' theory building or evaluation. Deep learning may be an impressive technological advance, but from the perspective of computational psychology, it can safely be ignored.

In light of our earlier conclusions, we can see that this reasoning is flawed. Multiple realization implies ontological autonomy, but it does not imply that lower-level details are irrelevant to higher-level inquiry. If an LLM replicates human linguistic abilities by implementing compositional syntactic structure, this will not motivate eliminating the constructs of syntactic theory, but it may help us *refine* or *revise* those theories. Which particular syntactic operations does the DNN seem to be employing (Manning et al., 2020)? What kind of compositionality does it implement (van Gelder, 1990)? Does the acquisition of this compositionality require hard-wired inductive biases, or can it be learned from the data over the course of training (Lake & Baroni, 2023; Russin, McGrath, Pavlick & Frank, Forthcoming; Russin, McGrath, Williams, Elber-Dorozko, Forthcoming)? Neural network models can help us to address each of these questions (McGrath et al., 2023). Just as with neuroscience, this will require interpretation and the careful extraction of functional characterizations of the network. Advances in this area are already underway (Manning et al., 2020) and hold real promise of identifying informative implementations.

Much ink has been spilled debating whether artificial neural networks are implementation-level models or offer competing cognitive or algorithmic accounts (Blank, 2023; Broadbent, 1985; Rumelhart & McClelland, 1985; Smolensky, 1988). The assumption is that the answer to this question will determine whether these networks have anything to teach us. This is misguided, because *not all implementation is 'mere.'* Even if, upon interpretation, LLMs seem to be implementing syntactic structure, they can still show us that these constructs are different than we took them to be and help us to explain their role in cognition, without explaining them away.

Fodor and Pylyshyn's discussion obscures the possibility of informative implementation by exploiting a certain mismatch between a strict conception of 'implementation' and the looser way that the notion gets employed in practice. On a strict application, we might require a full specification of the computations to be implemented in advance, so as to map each of these formal state transitions onto physical states at the implementation level (Chalmers, 1994). This strict notion of implementation is the one that is applicable to chess, for example, and it leaves little room for variation—any divergence from the pre-specified rules would make it a different game.

In the practice of cognitive science, however, this kind of fully specified functional description is the exception, not the rule. For the most part, the psychological constructs that modelers talk about implementing (e.g. compositionality, the Language of Thought, constructs like 'cognitive control' or 'executive function') are distinctly underspecified and leave room for a successful 'implementation' to diverge in illuminating ways from prior theoretical assumptions, enhancing our understanding of the capacity in question.

Some theorists would object to using the notion of 'implementation' in this less demanding sense and insist that a neural network model that diverges in these ways is an alternative cognitive account, rather than an informative implementation (Millière, 2024; Smolensky, 1988). As we see it, the problem arises not from this more inclusive use itself, but from equivocating between the two senses, as Fodor and Pylyshyn seem to—employing the more lenient notion to establish that a DNN counts as an implementation, then tacitly switching to the more demanding assumptions that would make this implementation 'mere,' thereby establishing a false sense of immunity from revision.

In our view, there is no harm in acquiescing to the more inclusive use of 'implementation,' *so long as we recognize that it does not license skepticism about the explanatory relevance of DNNs.*

The resulting position leaves room for both psychological theory and computational modeling to make important, mutually reinforcing contributions to cognitive science. Higher-level theory generates useful constructs, like executive function and cognitive control, and identifies key explanatory properties, like compositionality. Computational modelers attempt to implement these constructs in 'runnable' (Cao & Yamins, 2021) models capable of replicating the behaviors that these properties are meant to explain. What makes DNNs especially exciting from this perspective is that the modeler need not possess a fully specified solution in advance. Rather, they set up the network architecture, provide it with data, and allow it to identify its own optimal solution over the course of training, which interpreters can then scrutinize to extract new insights that may motivate revision to the original higher-level theory (McGrath et al., 2023).

Insofar as the mere implementation objection functions as a blinder to screen off this revisionary potential and insulate higher-level theory, it receives no support from the multiple realizability thesis and needs to be rejected.

# Bechtel, Mundale and the Revival of Mind-Brain Identity Theory

Severing the connection between multiple realizability and methodological autonomy prevents proponents of the former from denying the theoretical significance of neuroscience and neural network modeling for psychology and linguistics, as we have just seen. It also undermines the contrapositive argument: it prevents those who reject methodological autonomy from denying the multiple realizability thesis on that basis. To illustrate the point, we will examine the influential attack on multiple realization launched in Bechtel and Mundale's (1999) paper, "Multiple Realizability Revisited: Linking Cognitive and Neural States." First, we raise an objection to their strategy for rejecting purported cases of multiple realization. Next, we show how severing the connection between multiple realization and methodological autonomy undermines their central argument.

Though the multiple realization thesis remains a source of strikingly wide-ranging philosophical agreement, the twenty-first century has seen a variety of new attacks (Bickle, 2010; Bickle, 2020). Bechtel and Mundale (1999), among other key articles (Shapiro, 2000), played a crucial role in motivating this second wave of debate.

Bechtel and Mundale's paper is rich with historical case studies and provides a detailed account of how knowledge of the brain has inspired novel cognitive hypotheses about visual information processing. Taken together, these cases build a compelling argument against methodological autonomy. However, their real goal is to undermine the multiple realization thesis and revive mind-brain identity theory. To this end, they outline a general strategy for dismissing purported cases of multiple realization.

The reason that multiple realization seems intuitive, Bechtel and Mundale argue, is that philosophers employ an overly simplistic notion of the brain states with which psychological states are to be identified. Following Putnam, 'brain states' are assumed to be individuated based purely on their physio-chemical properties. Bechtel and Mundale call this a "philosopher's fiction" (Bechtel & Mundale, 1999). They argue that functional properties also play an essential role in neuroanatomy and the division of the brain into distinct areas. This opens the door to a more sophisticated notion of a 'brain state,' which they argue will be *shared* between individuals and cross-species, despite physical differences. Other creatures may be capable of experiencing pain or forming beliefs, as proponents of multiple realization claim, but when this holds it will be because they share the same brain states as well. As such, Bechtel and Mundale conclude that multiple realization is unwarranted and mind-brain identity can be preserved.

This argumentative strategy may look plausible when the only alternate realizations at hand are biological organisms, but consideration of artificial implementations, as might be found in DNNs, reveals its limits.

It is one thing to point out that brain mapping is not completely independent of functional considerations; it would be another thing entirely to stretch the notion of a 'brain area' so far as to include non-biological systems, like artificial neural networks. To do so is, in effect, to make the individuation of brain areas *solely* dependent on their function, rejecting anatomical considerations entirely. This constitutively functional notion of a brain area—on which *what it is* to be a prefrontal cortex, say, is to be involved in executive function—can serve as a useful shorthand for picking out functional analogues of human brain areas (Russin et al., 2020). But it is not a strong enough foundation to bear the weight of identity theory.

Moreover, even if we were to accept this extended, coarse-grained notion of a 'brain area,' it would not so much *repudiate* multiple realization as *recapitulate it at the neural level*. Executive function would no longer be multiply realized—it could be identified with activity in the PFC. But the PFC itself would be functionally individuated, substrate independent, and multiply realizable in everything from silicon circuits to alien substances. The victory for identity theory would thus be pyrrhic—it would preserve the identification of mental states with brain states only by transforming the notion of a 'brain state' into just the kind of abstract, functionally specified, multiply realizable state that Putnam and Fodor were advocating for from the outset.

The core problem with Bechtel and Mundale's paper, however—and our central reason for focusing on it here—is the structure of its central argument, which attempts to transform their objections to methodological autonomy into an argument against multiple realization. The argument has the following structure: (1) If mental states are multiply realized, then higher-level cognitive science is autonomous, and neuroscience is irrelevant. (2) A careful assessment of the science shows that neuroscience is not irrelevant—it has contributed a wide array of important insights to cognitive theory. Therefore, (3) the multiple realization thesis is false.

This argument turns the Fodorian case for methodological autonomy on its head. It accepts the validity of the inference captured in conditional premise (1), but argues that the falsity of its consequent means we must "jettison the multiple realizability thesis" (Bechtel & Mundale, 1999).

Bechtel and Mundale's attack on methodological autonomy is convincing, but this argument fails for the same reason as Fodor and Pylyshyn's: premise (1) is false. Multiple realization may entail ontological autonomy, but it does not entail methodological autonomy.

Severing this inferential link thus cuts in both directions. It undermines the Fodorian attempt to insulate higher-level inquiry in psychology and linguistics. But it also means that we cannot leap from the fact that neuroscience is relevant to the discovery of psychological truths to the claim that psychological states *just are* brain states. The brain can teach us about the structure of the mind, even if the mind may have an indefinite number of alternate realizations.

In the age of deep learning, this lesson is more salient than ever. Contemporary DNNs are the most plausible examples to date of alternate realizations of human psychological capacities, from the auditory recognition of speech and music (Kell et al., 2018), to natural language processing (Schrimpf et al., 2021), and competence with syntactic island constraints (Wilcox et al., 2023; though see Lan et al., 2023). Even in Bechtel and Mundale's central case of visual processing, DNNs have taken inspiration from the hierarchical organization of the brain's visual stream to replicate key capacities like object recognition and multi-object tracking (Zhao & Xu, 2019).

Difficult questions remain about the exact nature of these capacities, but it is plausible that today's models or their successors will provide tangible realizations of shared functional states in non-biological systems, just as Putnam predicted. Consideration of these recent findings offers a striking reaffirmation of our central claim: the fact that studying the brain is a crucial path towards understanding *and implementing* higher-level cognitive processes need not undermine a commitment to the possibility of multiple realization (Marblestone, Wayne & Kording 2016; Botvinick et al., 2019).

## Discussion

Let's take stock. We have argued that, even if mental states are multiply realized, this does not mean that cognitive scientists can afford to ignore neuroscience and neural network modeling. And we have explored key upshots of rejecting this inference from multiple realization to methodological autonomy for mind-brain identity theory and for Fodor and Pylyshyn's attempt to insulate higher-order cognitive theory.

It is important to be clear about the limits of these observations. Severing this inferential connection does not settle the issue of multiple realization and mind-brain identity theory on its own. Bechtel and Mundale's defense of mind-brain identity theory is flawed, but the debates that it incited are ongoing and fellow opponents of multiple realization have developed arguments based on different and even opposed premises (Cao, 2022). Nor does severing this connection demonstrate that lower-level details discovered by neuroscience or through computational modeling *must* be integrated into higher-level cognitive inquiry. Though methodological autonomy is not entailed by the multiple realizability thesis, there could be other considerations that motivate a strictly top-down approach to the study of human cognition. To conclude, we would like to emphasize further ramifications of the deep learning revolution for both issues.

Regarding mind-brain identity theory, it is fair to say that the rise of DNNs poses a new challenge—or rather, reinstates an old one. As noted above, the shift from discussion of multiple realizability to multiple realization has made it easier for defenders of identity theory to cordon off the sorts of speculative thought experiments and far-flung metaphysical possibilities that prove problematic for the view. As artificial intelligence shifts from the realm of science fiction into our everyday reality, this clean separation begins to look less feasible.

Conclusively demonstrating that a specific neural network model constitutes an alternative realization of a human cognitive capacity will require a more detailed technical discussion, as well as further engagement with subtle philosophical issues. This is an important topic for future work at the cross-roads of philosophy, cognitive science, and artificial intelligence. Here, we may simply conclude that the enterprise of *engineering* intelligent systems, as opposed to studying those found in nature, has finally attained the kind of breakthrough that makes the multiple realization of sophisticated cognitive capacities seem like a viable practical aspiration.

Regarding the methodological autonomy of cognitive science, we would like to close by suggesting that the success of deep learning may ultimately present a stronger challenge than has yet been discussed. Our focus here has been on establishing that DNNs *can* be relevant to higher-level cognitive theories. A more ambitious thesis is that they *must* be—that top-down approaches, which attempt to formulate cognitive-level explanations of human intelligence without considering implementation-level details, are doomed to failure. The rise of deep learning may furnish the resources for an argument in favor of this more radical conclusion.

The main methodological upshot of the deep learning revolution is what Richard Sutton has called the "bitter lesson" (Sutton, 2019). The history of AI suggests that progress towards engineering alternate realizations of psychological capacities depends not on thinking up clever theories that human modelers might try to implement, but on simply scaling the models up (both in terms of their computational power and the amount of data that they have access to) and allowing them to search through the solution space themselves. In the long run, Sutton writes, "building in how we think we think does not work" (Sutton, 2019). The theoretical solutions that researchers formulate may simply be too brittle or idealized to adequately capture the "tremendously, irredeemably complex" integrated functions involved in human cognition (Sutton, 2019).

What makes this lesson 'bitter' is that it undermines the classical conception of cognitive scientists as software reverse-engineers—or at least suggests that we may not be capable of carrying out the job on our own. There are differences between engineering a model to reproduce a capacity and providing an explanation of that capacity, and these differences might allow one to resist generalizing this bitter lesson. For example, it may be that the kinds of idealizations that have prevented traditional linguistic theories from yielding 'runnable' cognitive models (Cao & Yamins, 2021) are nevertheless necessary or useful from an explanatory perspective, providing succinct, intelligible descriptions that would not be discovered simply by adding more compute to a neural network.

If this bitter lesson *does* prove to be a more general moral, however, it will signal the end for traditional top-down approaches to cognitive science, as well as the necessity of closer integration with neural network modeling. For now, these remain speculative possibilities and open questions for future work. To conclude, we reiterate that, with the rise of deep learning, the multiple realizability thesis has renewed philosophical and empirical significance.

## Acknowledgments

We are deeply grateful to Roman Feiman and Ellie Pavlick for many hours of close collaboration on related issues and for helping us to create the kind of interdisciplinary research group that we believe to be crucial to progress in this area. We would also like to thank Riki Heck, Chris Hill, Elizabeth Miller, Josh Schechter, and our anonymous reviewers for helpful comments on an earlier draft of this paper.